\documentclass[10pt,twocolumn,letterpaper]{article}

\usepackage{cvpr}              

\usepackage{graphicx}
\usepackage[accsupp]{axessibility}
\usepackage{amsmath}
\usepackage{amssymb}
\usepackage{booktabs}
\usepackage[export]{adjustbox}
\newcommand\blfootnote[1]{%
  \begingroup
  \renewcommand\thefootnote{}\footnote{#1}%
  \addtocounter{footnote}{-1}%
  \endgroup
}

\usepackage[pagebackref,breaklinks,colorlinks]{hyperref}

\usepackage[capitalize]{cleveref}
\crefname{section}{Sec.}{Secs.}
\Crefname{section}{Section}{Sections}
\Crefname{table}{Table}{Tables}
\crefname{table}{Tab.}{Tabs.}

\def\cvprPaperID{00030} 
\def\confName{CVPR}
\def\confYear{2022}

\begin{document}

\title{Guiding Attention using Partial-Order Relationships for Image Captioning}

\author{Murad Popattia$^{1}$, Muhammad Rafi$^{1}$, Rizwan Qureshi$^{1, 2}$, Shah Nawaz$^{3\dagger}$ 
\\
$^{1}$National University of Computer and Emerging Sciences, Karachi, Pakistan, \\ 
$^{2}$Hamad Bin Khalifa University, Doha, Qatar \\
$^{3}$Pattern Analysis \& Computer Vision (PAVIS) - Istituto Italiano di Tecnologia (IIT) \\
{\tt\small muradmansoor189@gmail.com},
{\tt\small muhammad.rafi@nu.edu.pk},
{\tt\small riahmed@hbku.edu.qa}, \\
{\tt\small shah.nawaz@iit.it}
}

\maketitle
\begin{abstract}
The use of attention models for automated image captioning has enabled many systems to produce accurate and meaningful descriptions for images. Over the years, many novel approaches have been proposed to enhance the attention process using different feature representations. In this paper, we extend this approach by creating a guided attention network mechanism, that exploits the relationship between the visual scene and text-descriptions using spatial features from the image, high-level information from the topics, and temporal context from caption generation, which are embedded together in an ordered embedding space. A pairwise ranking objective is used for training this embedding space which allows similar images, topics and captions in the shared semantic space to maintain a partial order in the visual-semantic hierarchy and hence, helps the model to produce more visually accurate captions. The experimental results based on MSCOCO dataset shows the competitiveness of our approach, with many state-of-the-art models on various evaluation metrics.
\end{abstract}

\section{Introduction}
\blfootnote{$\dagger$ Current Affiliation: Deutsches Elektronen-Synchrotron (DESY)}
\blfootnote{Email: shah.nawaz@desy.de}
Recent success of deep neural networks in computer vision, speech, and natural language processing have prompted academics to think beyond these fields as separate entities, instead solving challenges at their intersections~\cite{hori2017attention, zhou2020unified,gallo2017multimodal,arshad2019aiding,nawaz2019cross,saeed2021fusion}. Generating descriptive and meaningful captions for images, and to capture its semantic meaning, is one such multimodal inference problem~\cite{hossain2019comprehensive, bai2018survey}. Despite its complexity, it has various applications, including visually- impaired assistance, intelligent chat-bots, medical report generation, self- driving cars, and many more~\cite{srivastava2018survey}. In general, an image captioning model should be able to find objects, their positions, map the relationship, as well as express this relationships in a human understandable language.

A typical image caption system consists of a convolutional neural network (CNN) and a recurrent neural network (RNN), with CNN as the image encoder and RNN as the sentence decoder~\cite{vinyals2015show, soh2016learning}. However, in order to capture the spatial context from the image in an efficient manner, other approaches such as \cite{yu2018topic, herdade2020image, 8451083, yao2016boosting} incorporate high-level information from topics or detected objects as semantic features to the decoder model. Another line of research was to make use of cross-modal associations between image and text features in a joint-embedding space. Earlier research work~\cite{kiros2014unifying, karpathy2015deep} treated images and caption as a symmetric relationship by using Euclidean or cosine distances to gauge similarities between these two modalities. On the other hand, in~\cite{vendrov2016orderembeddings} treated these associations as asymmetric by enforcing a hierarchical order within the embedding space, and has shown to perform better than symmetric relationships.

\begin{figure}[t]
\centering \includegraphics[width=0.5\textwidth,center]{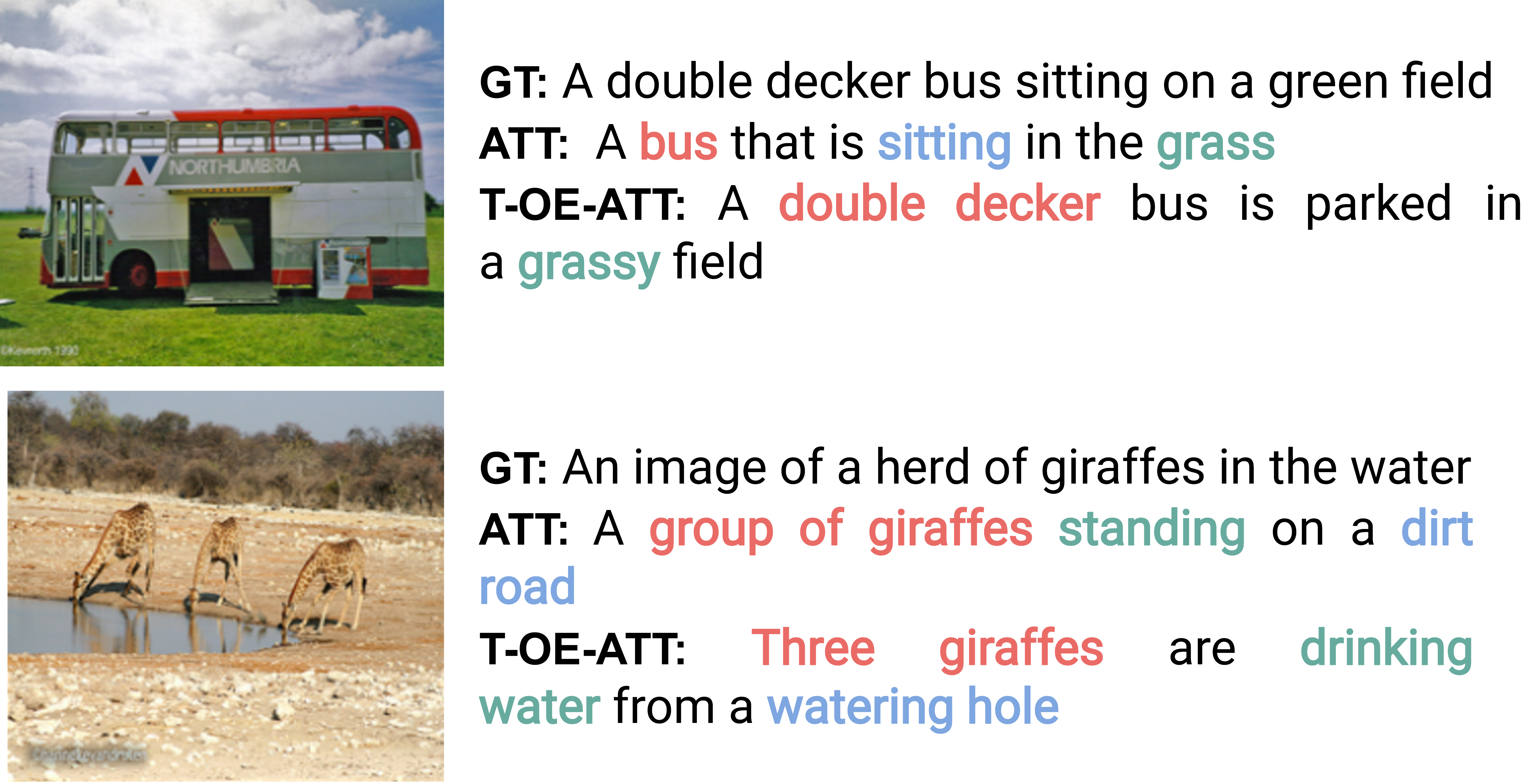}
\caption{Examples of generated captions by humans \textbf{(GT)}, attention \textbf{(ATT)} and using guided attention \textbf{(T-OE-ATT)}. The words higlighted in respective colors denote a comparison between the semantic detail captured by the approaches used.}
\vspace{-2mm}
\label{figure_qual}
\end{figure}

Further improvement in this framework, is the introduction of attention mechanism~\cite{spratling2004feedback}, which allows the decoder to focus on a sub-region of the image, when predicting the next word in the caption~\cite{xu2015show}. Despite of focusing only on spatial attention, Lu et al. ~\cite{lu2017knowing} presented a novel adaptive mechanism for helping the attention module to learn, when to shift between spatial and temporal context during word prediction. In addition, Anderson et al. \cite{anderson2018bottom} improves the attention process by first detecting a set of salient image regions (bottom-up) and then attending to these fixated regions (top-down).~\cite{Yao_2018_ECCV} builds upon this concept by exploiting the semantic relationships between the detected spatial regions using GCN (Graph Convolution Networks). \cite{DBLP:journals/corr/abs-1908-06954} also make use of a similar approach but instead modify the attention module by adding self-attention module on top of the conventional attention mechanism, which helps the decoder to draw relations between various attended vectors.

On the other hand, Jiang et al. \cite{jiang2018recurrent}, focused on increasing the semantic information fed to the decoder by using a fusion of multiple encoders, each focusing on a different view point, to build better representations for the decoder. Likewise, Wang et al. \cite{Wang_Chen_Hu_2019} also worked in a similar direction that guides attention using a hierarchy of semantic features. However, lack of inter-feature correlations between these encoders makes it difficult for the decoder to leverage the association from the resulting joint representations. Lastly, despite relying on spatial cues from encoded features, Ke et al. \cite{ke2019reflective} worked on improving the temporal coherence of words during descriptions by applying attention on both visual and textual domains.

Alongside the same line of work of incorporating semantic associations between different spatial regions using GCNs \cite{Yao_2018_ECCV}, our idea is to make use of multi-modal representations such as ordered embeddings \cite{vendrov2016orderembeddings} as our semantic feature vectors to guide the attention module. Similar to the late-fusion of features as done in  \cite{Yao_2018_ECCV}, we instead use a weighted summation as our fusion mechanism to fuse these embeddings.

Overall the main contributions of our work are three-fold: \begin{itemize}
    \item We make use of ordered embedding features for topics and images to guide the attention module instead of feeding them as low-level features. This step has been shown to improved metrics, see ablation study~\textit{(Section~\ref{subsec:abl})}.
    \item We incorporate a global weighted sum for fusing ``visual'' and ``temporal'' states instead of feeding them at each time-step separately which helps the model to learn the best estimation of the attention required for each image.
    \item Lastly, we present an ablation study of each contribution and how it effects the overall performance of the model on the MSCOCO dataset.
\end{itemize}

\begin{figure*}[t]
\centering \includegraphics[width=0.81\textwidth,center]{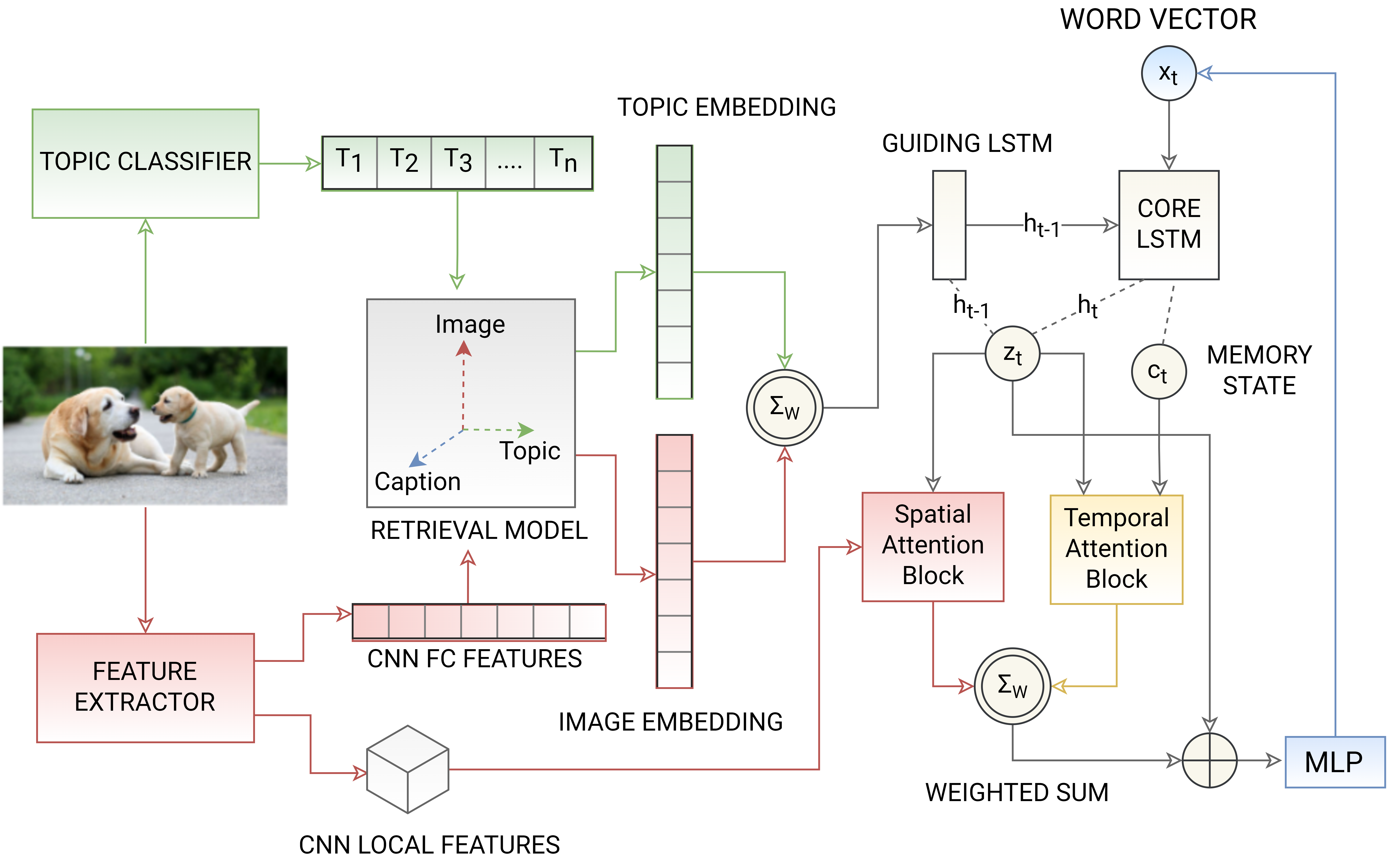}
\caption{The overall framework of the proposed model, where \(\Sigma_{w}\)\ represents a weighted-summation and \(\oplus\) denotes matrix addition. The model consists of a feature extractor and a topic classifier to extract spatial features and topics given in an image. These semantic attributes are then fed into a retrieval model which arranges an image, topic and caption triplet in a partial-order hierarchy. The resultant embeddings are then late-fused using weighted summation and then fed into a 'guiding LSTM'. The 'core-lstm' then makes use of this hidden state for temporal attention. Consequently, two separate attention blocks are used, each attending to different aspects of decoding and the resulting attention vectors are late-fused again in a weighted fashion to produce captions.}
\vspace{-2mm}
\label{figure}
\end{figure*}

\section{Methodology}

\subsection{Overall Framework}
Our approach follows the traditional encoder-decoder framework, where the encoder is responsible to pass on features used by the decoder to output the most likely word during captioning. Figure \ref{figure} illustrates the overall framework. 

Similar to recent approaches of sending objects or topics during encoding~\cite{8451083, yang2017image}, we used topics instead of objects to capture both the "actors" as well the ``activities'' binding them. The encoder consists of three components: \textbf{1)} topic classifier \textbf{2)} feature extractor and the \textbf{3)} retrieval model. We use a pre-trained deep CNN model as a feature extractor to extract visual features from the image and train a multi-label topic classifier to predict topics for given images. After that, we train a retrieval model which embeds captions, image and topics into a shared semantic space, in which a similarity score can be calculated between them. Interestingly, using embeddings helps to better learn the latent relationships between image and topic features, lost during feature extraction. This helps the attention module in describing and discriminating spatial regions more effectively. \textit{(Details in Section~\ref{subsec:retrievalmodel})}

Inspired from the simple yet effective architecture defined in \cite{anderson2018bottom}, we used two LSTM branches in the decoder i.e. the \textit{guiding-lstm} and the \textit{core-lstm}. Here, we define a weighted sum of the semantic embeddings of both the images and topics, as input to the \textit{guiding-lstm} at the first time-step, which gives the model a better understanding of the alignment of visual features and the topics. We then utilize its hidden state \textit{h\textsubscript{t-1}} for guiding the language LSTM and the context vector \textit{z\textsubscript{t}} used for attention. Similar to using a visual sentinel \cite{lu2017knowing}, we used a weighted summation for fusing the attention weights instead of a sentinel gate to shift between spatial and temporal attentions. This allows for a more simpler architecture in terms of learning parameters involved, whilst maintaining the accuracy during word prediction.

\subsection{Topic Classifier}
\label{sec:tc}
For extracting topics~\textit{T}, the ground-truth captions are concatenated to form documents \textit{D}, where each document \textit{d} corresponds to captions \textit{C}, for a given image and contains a set of words \textit{W}. After that, we train a Latent Dirichlet Allocation (LDA) model~\cite{blei2003latent}, which is a probabilistic model to learn the topic representations from documents. The trained topic model outputs a set of topic probabilities \textit{\textbf{T}=\{T\textsubscript{1}, T\textsubscript{2}, T\textsubscript{3}, ... T\textsubscript{n}\}}.

For training the classifier, the topics are sampled and converted to one-hot representations using the following function:

\begin{equation} \label{eq:teq}
f\textsubscript{t\textsubscript{i} \(\subseteq\) T\textsubscript{i}}(x) = \left\{ \begin{array}{rcl}
1 & \mbox{if} & P(x)\geq0.1 
\\ 0 & \mbox{else}
\end{array}\right.
\end{equation}

where \textit{t\textsubscript{i}} represents a single topic from a set of topics \textit{T}, for image \textit{i} from a set of images \textit{I} and \textit{P(x)} represents the topic-confidence from LDA. We formulate this as a multi-label classification problem, since an image can have multiple topics. A pre-trained CNN model is used to extract image features which are then fed into a feed-forward neural network with a sigmoid activation for the prediction layer. This layer outputs an \textit{(N\textsubscript{i}\(\times\)N\textsubscript{t})} vector where \textit{N\textsubscript{i}} corresponds to the number of images and \textit{T\textsubscript{t}} are the number of topics. We report the evaluation for the topic classifier in \textit{Section~\ref{subsec:tc}} of the paper.

\subsection{Retrieval Model}
\label{subsec:retrievalmodel}
The architecture of the retrieval model is inspired by the approaches in~\cite{yu2018topic, vendrov2016orderembeddings}. It follows the idea of \cite{karpathy2015deep} to align caption and images in the same space, but with a partial-order relation rather than a symmetric relation. This is a more intuitive approach as images have captions with different levels of details, and because the captions are so dissimilar, it is impossible to map both their embeddings close to the same image embedding using a symmetric distance measure like cosine similarity. Nevertheless, maintaining order is robust to such affect, as dissimilar caption can have embeddings placed very far away from the image, while remaining above it in the partial order. The partial order relation can be defined as: \\ \textit{ \(x \preceq y \) \(\Longleftrightarrow \forall x \forall y (x\geq y)\)}. This imposes for all values of the vector \textit{x} to be greater than all values of the vector \textit{y} in the embedding space to maintain order. 

We start with three entities i.e. images \textit{I}, topics \textit{T} and captions \textit{C}. As per \cite{vendrov2016orderembeddings}, we utilized domain-specific encoders to extract features for training the embeddings. For images and topics, we utilized the fully-connected features from the feature-extractor and the topic features from the topic classifier respectively. While for captions, we used a Gated Relu Unit (GRU) as the RNN based text-encoder instead of an LSTM, because of its computational efficiency. These feature vectors are then weighted with W\textsubscript{I}, W\textsubscript{T} and W\textsubscript{C} before being projected in the embedding space:

\begin{equation} \label{eq1}
\vspace{1mm}
O\textsubscript{i} = \|W\textsubscript{I}\cdot f\textsubscript{FE}(I)\|^{2}
\end{equation}
\begin{equation} \label{eq1}
\vspace{1mm}
O\textsubscript{t} = \|W\textsubscript{T}\cdot f\textsubscript{TC}(T)\|^{2}
\end{equation}
\begin{equation} \label{eq1}
O\textsubscript{c} = \|W\textsubscript{C}\cdot GRU(C)\|^{2}
\end{equation}

 O\textsubscript{i}, O\textsubscript{t}, O\textsubscript{c} represents the order embeddings of image, topics, and captions respectively. \textit{f\textsubscript{FE}(I)} represents the image features from the feature-extractor, while~\textit{f\textsubscript{TC}(T)} represents the features from the topics classifier. We use L2-Norm during encoding instead of an absolute value function to mitigate overfitting~\cite{vendrov2016orderembeddings}. 

\textbf{Similarity Function} The general notion of similarity between two vectors \textit{x} and \textit{y} in the embedding space can hence be quantified as the degree to which a pair of points violates the partial order \textit{ \(x \preceq y \)} \cite{vendrov2016orderembeddings}:
\begin{equation} \label{eq1}
    S(x, y) = - (\|max(0,O\textsubscript{y}-O\textsubscript{x})\|^{2})
\end{equation}

where \textit{O\textsubscript{x}} and \textit{O\textsubscript{y}} represents the encoded feature vector in the embedding space. The negative sign constitutes to the fact that a positive difference between \textit{O\textsubscript{y}} and \textit{O\textsubscript{x}} denotes violation of the order penalty.

\textbf{Loss Function} As previous works which learn embedding in cross-modal retrieval tasks \cite{kiros2014unifying, karpathy2015deep}, we re-use the pair-wise ranking loss objective to increase the similarity for the matching pairs and vice-versa for the contrastive terms by a margin \(\alpha\):
\vspace{-1mm}
\begin{equation} \label{eq1}
\begin{split}
    L(x, y) = \sum_{(x,y)}(\sum_{x'}max\{0, \alpha - S(x, y) + S(x', y)\} + \\
    \sum_{y'}max\{0, \alpha - S(x, y) + S(x, y')\})
\end{split}
\end{equation}

where \((x,y)\) is the ground-truth pair while \((x',y)\) and \((x,y')\) are constrastive terms. Our hierarchy has image at the top of the partial order, followed by captions which are then bounded by the topics. Hence, the total loss can be defined as the summation of losses over all three partial orders:
\begin{equation}
    L = L(I, C) + L(I, T) + L(C,T)
\end{equation}


\subsection{Caption Generation}
\label{subsec:cg}
We now describe the decoding phase of the model. The trained encoding functions \(O\textsubscript{i}\) and  \(O\textsubscript{t}\) are used to produce relevant embeddings for image and topics during feature extraction. We then used a weighted-summation ( \(\Sigma_{w}\) ), of these embeddings:

\begin{equation}
    \Sigma_{w(OE)} = \lambda \cdot O_{i} + (1-\lambda) \cdot O_{t}
\end{equation}

where \(\lambda\) is a learnable parameter. The reason for a weighted-sum is to allow the model to learn the relative importance of each embedding during training. Different from the approach of \cite{yu2018topic}, we focused on guiding the decoder in a \textit{3-way} manner i.e. using the embedding information, visual features and reliance on past information from the hidden states.

\textbf{Dual-LSTM branch} We used an auxiliary \textit{guiding-lstm}, to process the information from the learned embeddings and feeding the hidden state information to both the \textit{attention} vector \textit{\(z_{t}\)} and the \textit{core-lstm} at initial timestep t = -1:

\begin{equation}
    h_{t-1} = LSTM_{g}(\Sigma_{w(OE)})
\end{equation}
\begin{equation}
    z_{t} = W_{g}h_{t-1} + W_{c}h_{t}
\end{equation}
\begin{equation}
    h_{t} = LSTM_{c}(x_{t}, h_{t-1})
\end{equation}

where \(h_{t-1}\) and \(h_{t}\) represents the hidden states at relevant timesteps, \(W_{g}\) and \(W_{c}\) are learnable parameters in the context vector \(z_{t}\). \(LSTM_{g}\) and \(LSTM_{c}\) represent the guiding and core LSTMs respectively. The initial hidden state for \(LSTM_{g}\) is essentially zeroes and hence not shown in the formulation. 

\textbf{Spatial Attention Block} This block is responsible to generate the attention distribution vector over the important visual regions of the image. Similar to the idea of~\textit{soft-attention}~\cite{lu2017knowing}, we utilize the context-vector \(z_{t}\) from equation 10 instead of just the hidden state information done in \cite{lu2017knowing}, in order to guide attention pertaining to the partial-order relation between the image and topic:
    
\begin{equation}
    \alpha_{t} = softmax(W_{\alpha}[W_{f}F_{L} + W_{z}z_{t}])
\end{equation}
\begin{equation}
    \rho_{s} = \sum_{i=1}^{N}\alpha_{ti}f_{i}
\end{equation}

where \(F_{L} = \{f_{1}, f_{2}, .... f_{N}\}\) represent the local image features from the convolution layer just before the FC layer of the feature extractor, \(\alpha_{t}\) denotes the attention weights over the features in \(F_{L}\), \(\alpha_{ti}\) denotes the weight over the i\textsuperscript{th} part of \(F_{L}\) and \(\rho_{s}\) denotes the \textit{spatial-context} vector.

\textbf{Temporal Attention Block} The temporal block guides the attention module whether the information is required at all, or the next word can be predicted using the past information stored within the decoder~\cite{lu2017knowing}. Likewise, we utilize the information from the LSTM's memory cell along with the context vector \(z_{t}\) which contains the residual embedding information from the previous timestep. It helps the temporal block decide whether the current timestep requires attending to visual features or not. This is illustrated below:

\begin{equation}
    \rho_{t} = \tanh(c_t) \bigodot \sigma({W}_{x}x_{t}+ W_{z'}z_{t}) 
\end{equation}

where \(c_{t}\) is the memory cell of the \textit{core-lstm}, \(x_{t}\) is the word vector at timestep \textit{t}, \(z_{t}\) denotes the context vector, \(\bigodot\) refers to an element-wise product and \(\rho_{t}\) denotes the \textit{temporal-context} vector.

\textbf{Word Prediction} Instead of keeping track of the temporal information for each word, we let the model generalize the ratio between these attentions using a weighted-summation ( \(\Sigma_{w}\) ). This is because ideally it is a more simpler approach to rely less on the attention gate at each timestep and generalize from the embedding context obtained from \(z_{t}\).
\begin{equation}
    \Sigma_{w(ATT)} = \mu \cdot \rho_{s} + (1-\mu) \cdot \rho_{t}
\end{equation}
We then calculate the word probability over a vocabulary of possible words at time t:
\begin{equation}
    p_{t} = softmax ( f_{MLP} ( z_{t} + \Sigma_{w(ATT)} ) )
\end{equation}

where \(f_{MLP}\) denotes a dense layer with ReLU activation.

\section{Experiments}

\subsection{Implementation Details}

As our model is divided into sub-components, we train each part separately instead of training them end-to-end.

\textbf{Feature Extractor} We use a ResNet-$152$~\cite{he2016deep} model trained on ImageNet dataset. The FC features are taken from the last layer of the CNN which have a dimension of $2048$\(\times\)$1$. We use  \(F_{L} = \{f_{1}, f_{2}, .... f_{N}\}, f_{i} \in R^{512}\) to represent the spatial CNN features at each of the \textit{N} grid locations where \textit{N = $49$}.

\textbf{Topic Classifier} For the training the topic model, we limit our vocabulary to top $5000$ and train the LDA on these features for $100$ iterations. We empirically set the number of topics to be $80$ for our case. Increasing the topics made the topic vectors more sparse and decreased the recall for the topic classifier. For the topic classifier, we used the image features \(R^{2048\times1}\) to be fed into a $5$-layer feed-forward NN, with the prediction layer \(R^{80}\)  having a sigmoid activation. The classifier was optimized using SGD with a learning-rate of $0.1$ and momentum $0.9$. The learning-rate was changed in case of plateauing with a patience of $0.4$ and a factor of $0.2$. 

\textbf{Retrieval Model} For the retrieval model, we reused the FC image features \(R^{2048\times1}\) from the feature extractor and the \(R^{80}\) topic features from the topic classifier in \textit{Section~\ref{sec:tc}}. The dimensions of the embedding space and the GRU hidden state in equation (4) were set to $1024$, and the margin $\alpha$ is set to $0.05$ as per \cite{vendrov2016orderembeddings}. 

\textbf{Caption Model} For the decoder, our model used LSTMs. The \textit{guiding} and \textit{core} LSTMs both have a dimension of $512$. For the captions, we use a word embedding size of $256$. During training, we see that downsizing and concatenating FC image features with this embedding improved results. The initial value for \(\lambda\) and \(\mu\) in equation (8) is set to $0.5$ for both, and learned during training. Furthermore, the number of units for \(f_{MLP}\) was set to $1024$. Lastly, for sampling the captions, we use beam size of $1$.  The whole model was optimized using Adam optimizer with a mini-batch size of 128 and learning rate of 0.001. The model trained for $10$ epochs on a Tesla T$4$ GPU and the training finished in $10$ hours to produce the results.

\subsection{Datasets}

We conducted experiments on the popular benchmark: Microsoft COCO dataset \footnote{https://cocodataset.org/} as this has been widely used for benchmarking in the related literature. Also, we adopt the ‘Karpathy’ splits setting \cite{ke2019reflective}, which includes 118,287 training images, and 5K testing images for evaluation. Some images had more than 5 corresponding captions, the excess of which are discarded for consistency. We directly use the publicly available code \footnote{https://github.com/tylin/coco-caption} provided by Microsoft for result evaluation, which includes BLEU, METEOR, ROUGE-L and CIDEr.

\begin{table*}[t]
\centering
\begin{tabular}{|lccccccc| } 
 \hline
  \textbf{Approaches} & \textbf{BLEU-I} & \textbf{BLEU-II} & \textbf{BLEU-III} & \textbf{BLEU-IV} & \textbf{METEOR}& \textbf{ROUGE-L} & \textbf{CIDEr}\\ 
 \hline \hline
 \textbf{Adaptive ATT} ~\cite{lu2017knowing}  &74.2 &58.0 &43.9 &33.2 &26.6 &-&108.5\\ 
 \textbf{LSTM-A} ~\cite{yao2016boosting} & 75.4 & -& -& 35.2& 26.9& 55.8& 108.8\\
 \textbf{RF-Net} ~\cite{jiang2018recurrent}  & 76.4 & 60.4& 46.6& 35.8& 27.4& 56.5& 112.5 \\ \textbf{Up-Down ATT} ~\cite{anderson2018bottom}  & 77.2  &	-&	-&	36.2&	27.0&	56.4& 113.5\\
 \textbf{HAN} ~\cite{Wang_Chen_Hu_2019} & 77.2 & 61.2& 47.7& 36.2& 27.5& 56.6& 114.8\\
 \textbf{RDN} ~\cite{ke2019reflective} & 77.5 & 61.8 & 47.9 & 36.8&  27.2& 56.8 & 115.3\\
 \textbf{GCN-LSTM} ~\cite{Yao_2018_ECCV} & 77.4 & -& -& 37.1& 28.1& 57.2& 117.1\\
 \textbf{AoA-Net} ~\cite{DBLP:journals/corr/abs-1908-06954} & 77.4 & -& -& \textbf{37.2}& \textbf{28.4}& \textbf{57.5}& \textbf{119.8}\\
 \specialrule{.1em}{.05em}{.05em}
 \textbf{Ours (T-OE-ATT) }&77.0&	61.2&	47.1&	35.9& \textbf{28.4}&	57.3& 115.9 \\
 \hline
\end{tabular}
\caption{Performance comparison on MSCOCO 'Karpathy' test split trained on a single-model using cross-entropy loss without CIDEr optimization. (-) indicates metrics not provided. All values are provided in percentages (\%) with the highest bold-faced.} \label{tab:cap1}
\end{table*}

\begin{table}[h]
\begin{center}
\begin{tabular}{|c|c|c|c|c| } 
 \hline
  \textbf{Approach} & \textbf{B-IV} & \textbf{METEOR}& \textbf{ROUGE-L} & \textbf{CIDEr} \\
 \hline\hline
 \textbf{Topic} &25.5 &	22.9 &	50.1 & 80.2 \\ 
 \textbf{T-OE\textsubscript{(VGG)}}  &34.4 & 27.8 & 56.5 & 112.7\\
 \textbf{T-OE\textsubscript{(Resnet)}}&35.4&28.2&	57.0&	114.4\\ 
 \textbf{T-OE-ATT} & \textbf{35.9}&\textbf{28.4}&	\textbf{57.3}&\textbf{115.9}\\ 
 \hline
\end{tabular}
\end{center}
\vspace{-6mm}
\caption{Ablation study on MSCOCO 'Karpathy' test split.} \label{tab:cap}
\end{table}

\subsection{Evaluation}

\subsubsection{Ablation Study}
\label{subsec:abl}

To study the effects of guiding the attention module, we design an ablation experiment to assess the effect of \textbf{1)} using an embedding space \textbf{2)} using a different feature extractor and \textbf{3)}  using embedding along with attention as shown in \textbf{Table \ref{tab:cap}}. We see that the initial approach of feeding topics as low-level features performs poorly. A dramatic improvement was seen when using an embedding space in the process. This confirms the hypothesis that embeddings serve as a better auxiliary guidance for attention. We term this as \textbf{(T-OE)}. Moreover, we assess the model's performance on a less accurate feature extractor such as VGG-19 \cite{simonyan2014very} which only incurred as small change in the metrics signifying that trained embeddings are robust to changes in the feature extractor. Lastly, we incorporate attention in the process \textbf{(T-OE-ATT)} and guide them using the trained embeddings which shows an improved score in all metrics, signifying the importance of the embeddings to guide attention.

\subsubsection{Quantitative Evaluation}
\label{subsec:qe}
In \textbf{Table \ref{tab:cap1}}, we compare our proposed architecture with recent state-of-the-art models on the MSCOCO dataset that make use of LSTMs in their decoder architecture. For fair comparison, we report the scores for single model for each approach that use the same CNN backbone as ours (ResNet \cite{he2016deep}), without using ensembling and CIDEr optimizations. 

Our approach is able to outperform RF-Net \cite{jiang2018recurrent} and HAN \cite{Wang_Chen_Hu_2019} signifying that using partial order is more suitable for building joint multi-modal representations as compared to using domain-specific encoders alone. Moreover, incorporating attention with T-OE, as shown in Table \ref{tab:cap}, also helps us outperform RDN \cite{ke2019reflective} over notable metrics such as \textbf{METEOR, ROUGE-L} and \textbf{CIDEr} which show that orthogonal improvements to encoder or decoder alone are less susceptible to improvement as compared to jointly improving both the feature representations and the caption generation process. It is worth noting that compared to our architecture, RDN \cite{ke2019reflective} and RF-Net \cite{jiang2018recurrent} have a greater number of learning parameters (1.15B parameters \cite{ke2019reflective} for RDN), whilst our decoder contains comprises of only 29M parameters and yet is able to produce competitive results. Both GCN-LSTM \cite{Yao_2018_ECCV} and AoA-Net \cite{DBLP:journals/corr/abs-1908-06954} use Faster-RCNN as their feature encoder which is able to feed in region-level information while our model uses only the fully connected features from the ResNet backbone and is still competitive over \textbf{METEOR} and \textbf{ROUGE-L} scores. It should also be noted that AoA-Net \cite{DBLP:journals/corr/abs-1908-06954} leverage the use of self-attention mechanisms which have been used alongside transformers and are able to produce state-of-the-art results. On the contrary, our work can be extended to incorporate region level information alongside topics or to use a different attention mechanism to improve results and has not been explored in this study.

As our model uses LSTMs for caption generation, hence this comparison does not take into account transformer-based architectures \cite{cornia2020meshedmemory, vaswani2017attention}. Transformers are a different class of architecture as compared to LSTMs as they do not follow the auto-regressive nature of LSTMs and process the inputs in a parallel fashion \cite{vaswani2017attention} so incorporating partial-order embeddings alongside this class of architecture could also be a favourable research direction.

\subsubsection{Qualitative Evaluation}

We assess our model qualitatively as illustrated in Figure~\ref{figure_qual}. The baseline model is based on model's output based on \textit{topic and image features}, while the guided attention model is based on \textit{topic and image embeddings}. Without embeddings, we see that attention lack descriptiveness of the context associated with the visual features such as \textbf{double-decker}, \textbf{grassy}, \textbf{drinking} etc. 

\begin{figure}[h]
\centering \includegraphics[width=0.5\textwidth]{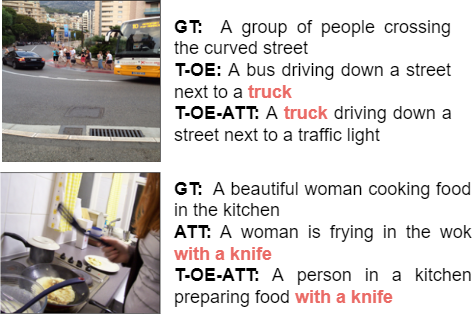}
\vspace{-6mm}
\caption{Examples of inaccurate captions from the model.}
\label{figure3}
\vspace{-3mm}
\end{figure}

We also see an influence when comparing ground truth captions where the model was able to capture semantic context like \textit{parked} instead of \textit{sitting} and \textit{drinking water} instead of \textit{in the water}. It is because the model is able to draw associations between objects and actions due to partial-order information from the underlying topics of the captions fed into the decoder module denoting how attention was guided.

However, as denoted in Figure~\ref{figure3}, the attention module can pick up on noise from these embedded features such as confusing between \textit{a bus} and \textit{a truck}. This is evident from \textbf{T-OE}, where the caption contains \textbf{truck} even though it is absent from the image. An explanation can be \textit{bus} and \textit{truck} being semantically closer in the embedding space. Moreover, relying on spatial attention can also lead to mis-classifying objects in the image from \textit{spatula} to \textit{knife}. This can be seen from the caption generated from the model without \textbf{T-OE} where the object is misidentified as a \textit{knife}.

\vspace{-2mm}

\section{Discussion}

\subsection{Evaluation of topic classifier and retrieval model}
\label{subsec:tc}

As the topic classifier and the embedding sub-space act as intermediaries to the final model, hence, we evaluate their performance on relative metrics as well. The output of the topic classifier is a set of sigmoid probabilities which are converted into one-hot encodings.  Using precision solely for evaluating one-hot encodings is not enough as we can see that a higher precision does not mean our model has a good recall.  Hence,  we use F1-score with a \(\beta\) more inclined towards recall. The highest F1-score was achieved in the COCO dataset which may be due to a larger amount of data being used to train the model. We summarize these results in \textbf{Table \ref{captopic}}.

\begin{table}[h]
\begin{center}
\begin{tabular}{|c|c|c|c|c| } 
 \hline
  \textbf{Dataset} & \textbf{Precision} & \textbf{Recall} & \textbf{F1-Score} \\
 \hline\hline
 \textbf{Flickr30k} & 60.08 & 42.33 & 43.56 \\ 
 \textbf{MSCOCO} & 77.54 & 60.48 & 61.52 \\ 
 \hline
\end{tabular}
\end{center}
\vspace{-6mm}
\caption{Performance results of the topic classifier on validation sets of Flick30k and MSCOCO dataset} \label{captopic}
\end{table}

For the order embedding model, we assess the quality of the model by treating it as a Caption Retrieval task.  The metric used in this experiment was Recall@K which refers to the percentage of recall achieved in top-k items. We summarize these results in \textbf{Table \ref{capemb}}.

\begin{table}[h]
\begin{center}
\begin{tabular}{|c|c|c|c|c|c|c|} 
 \hline
  \textbf{Dataset} & \textbf{R@1} & \textbf{R@5} & \textbf{R@10}\\
 \hline\hline
 \textbf{Flickr30k} & 35.2 & 61.9 & 73.4\\ 
 \textbf{MSCOCO} & 49.5 & 79.6 & 89.3\\ 
 \hline
\end{tabular}
\end{center}
\vspace{-6mm}
\caption{Performance results of the retrieval model on validation sets of Flick30k and MSCOCO dataset} \label{capemb}
\end{table}

Nevertheless, the scores for both the topic classifier and the retrieval model were not state-of-the-art but were enough to extract suitable features for the training images. Respective improvements to the models in terms of fine-tuning or using a different architecture, might positively impact the overall accuracy during captioning but is beyond the scope of this paper.

\subsection{Visualizing the embedding space}

\begin{figure}[t]
\centering \includegraphics[width=0.5\textwidth]{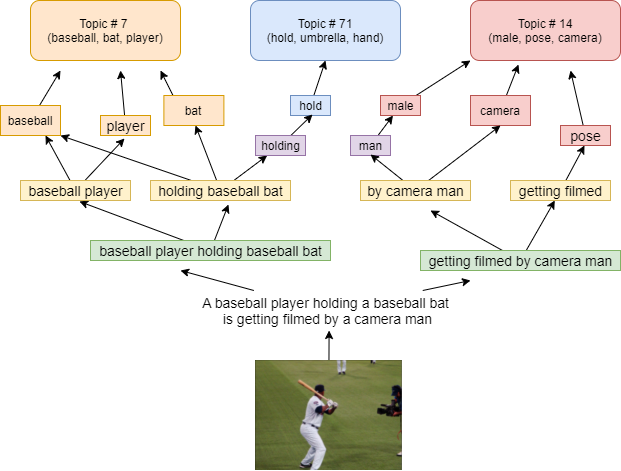}
\caption{Representation of order in the embedding space}
\label{embvisual}
\end{figure}

In this section, we present a high-level visualization of the partial-order structure between images, topics and captions in the embedding space, as shown in Figure~\ref{embvisual}.

The embedding space consist of three modalities, with images being at the highest order, captions being at the center and topics being at the lowest order of the hierarchy posing a lower bound for the captions. This hierarchical arrangement also conforms with the cognitive arrangement of these modalities. Images are generally abstract points from which we derive meaning about its context while separate words such as topics can be used to complement images but do not contribute to any meaning on their own. Captions on the other hand, describe a story which the spatial cues of the image support. 

We can then visualize these captions as a collection of words each of which can constitute to a topic. Treating the problem as a caption retrieval task, where given an image the model outputs the set of all possible captions, setting a lower bound with topics helps constraint this search space and helps reduce noise from overlapping caption regions. \cite{yu2018topic}

\subsection{Analysis of the weighted summation for attention}

\begin{figure}[t]
\centering \includegraphics[width=0.5\textwidth]{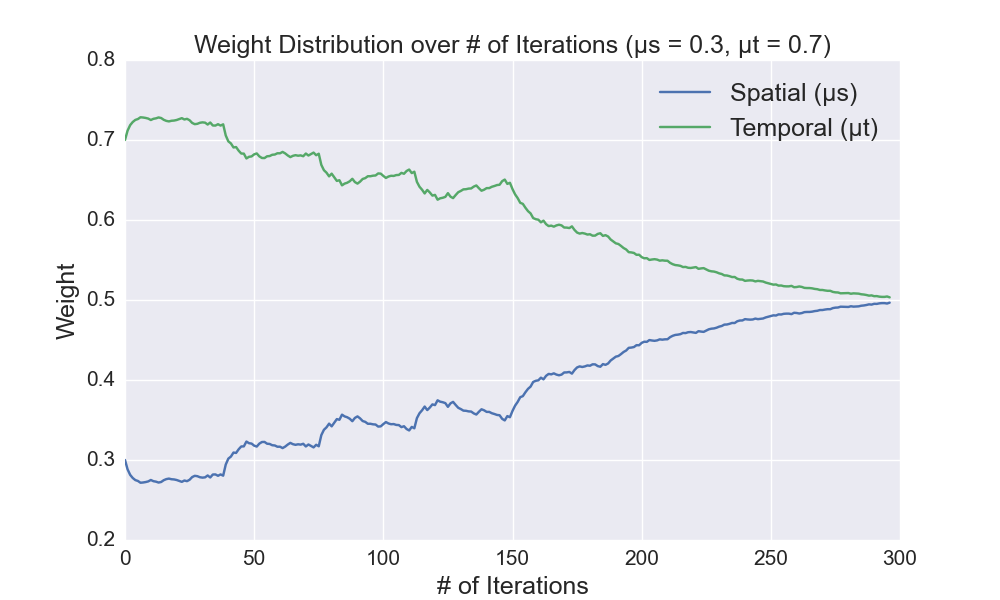}
\caption{Weight distributions of spatial and temporal attention for several iterations on MSCOCO dataset.}
\label{weightSum}
\end{figure}

Contrasting to the approach followed in \cite{lu2017knowing}, where the model is trained to shift attention at each word prediction step, we constraint the model in determining an overall ratio of the spatial or temporal attention needed for word prediction and keep this as a static value for all succeeding predictions. However, despite setting the values randomly, we allow the decoder to generalize from a set of caption on the amount of attention needed for each caption. For testing, we set the temporal context \(\mu\) to be 0.3 for spatial and consequently, 0.7 for temporal attention. The reason for a higher ratio for temporal context is because it complements the RNNs capability to work with sequences. For the model we use the \textbf{ATT} approach where the image features are fed directly as spatial cues to the decoder. \textbf{Figure~\ref{weightSum}} shows the learned ratios after several iteration of training.

\begin{figure}[t]
\centering \includegraphics[width=0.5\textwidth]{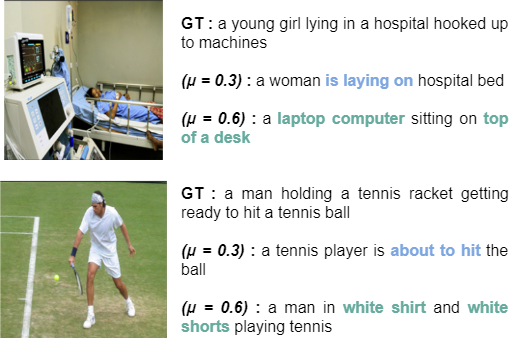}
\caption{Sampled captions on varying \(\mu\) values on COCO dataset. A higher value of \(\mu\) denotes more weight being given to the spatial flow of information within the decoder and viceversa.}
\label{muvalue}
\end{figure}

It can be seen that the model gradually learns to increase the gradient flow from the spatial block of the attention module, signifying the need of visual attention. However, we do notice some peaks for the flow of temporal information. A plausible reason is because while visual information is necessary, it may not always be inline with temporal coherence when describing images. Hence, we sample captions with different values for \(\mu\) as shown in Figure~\ref{muvalue}.

For a lower value of \(\mu\) in the formulation \textit{15} in \textit{Section~\ref{subsec:cg}}, the model allows the flow of temporal information in the decoder and hence we see a time-relative aspect in sentences with phrases such as \textit{"about to hit"} and \textit{"is laying on"}. On the contrary, if we shift the value of \(\mu\) higher, it boosts the gradient flow from the spatial block filling in visual details from the image such as \textit{"white shirt"}, \textit{"white shorts"}, \textit{"laptop computer"}, \textit{"top of desk"}. However, we see that despite being rich in scene-specific details, the model misses out the global context of the image imposing the need for a good balance between both the attention modules. This is the reason we allow the model to learn these weights during training.

\section{Limitations}

In this section, we discuss the architectural limitations to our work and also explore future extensions to this approach. Firstly, the performance of the decoder phase is dependant on the output from the topic classifier and bottle-necks the overall improvement from training. Moreover, most recent works such as GCN-LSTM \cite{Yao_2018_ECCV} and AoA-Net \cite{DBLP:journals/corr/abs-1908-06954} make use of Faster-RCNN to feed in region-level information and hence incorporating these object-level associations alongside topics in the multi-modal embedding space are susceptible to increase in efficacy of the approach used. Another limitation of our work is the use of traditional attention mechanisms. Our study make use of soft-attention mechanism which involves the averaging of feature maps. Comparing our approach with HAN \cite{Wang_Chen_Hu_2019} which also makes use of soft-attention mechanism, we gain a relative improvement as discussed in Section~\ref{subsec:qe}. However, our approach struggles against AoA-Net \cite{DBLP:journals/corr/abs-1908-06954} which uses a more robust attention mechanism. Moroever, the use of self-attention has been shown to improve performance over traditional attention mechanisms such as \cite{DBLP:journals/corr/abs-1908-06954}, more notably in transformers \cite{cornia2020meshedmemory, vaswani2017attention} and hence can be incorporated with the use of these multi-modal embeddings to improve performance. Lastly, using recent reinformcement learning based techniques such as CIDEr optimizations~\cite{DBLP:journals/corr/RennieMMRG16} have yielded state-of-the-art results for image captioning, incorporating them with our study may further boost the performance over the metrics used.

\section{Conclusion}

\par In this work, we proposed a new approach to guide the attention model by exploiting partial-order relationships between image, captions and topics. Arranging the image and textual modalities in an asymmetric fashion results in more effective learning of the latent space. Hence, we make use of a multi-modal embedding space that is able to arrange the visual and textual modalities in an asymmetrical hierarchy where the caption embeddings are bounded between image and topic features. We then make use of these joint representations to guide the attention module. An extensive ablation study was also performed to indicate that using ordered embeddings, the attention model was able to draw accurate links between semantically important regions of the image when attending to them, which helped improve the overall interpretability, syntax and descriptiveness of the captions. The proposed architecture was not only simpler in terms of complexity, but also competitive with many recent LSTM-based architectures. For next steps, a promising direction can be to incorporate the highlighted approach with transformers or leveraging the model architecture to be trained in an end-to-end manner.

{\small
\bibliographystyle{ieee_fullname}
\bibliography{egbib}
}
\end{document}